\documentclass[runningheads]{llncs}
\usepackage[T1]{fontenc}
\usepackage{graphicx}
\usepackage{amsmath,amssymb}
\usepackage{booktabs}
\usepackage{enumitem}
\usepackage{xcolor}
\usepackage{cite}
\usepackage{marvosym}
\usepackage{wrapfig}
\usepackage{float}
\definecolor{morandigrey}{RGB}{235,235,230}
\definecolor{morandired}{RGB}{228,212,210}
\definecolor{morandigreen}{RGB}{214,226,220}
\definecolor{morandiline}{RGB}{180,180,170}
\definecolor{morandiblue}{RGB}{196,206,216} % soft steel-blue
\definecolor{morandidark}{RGB}{90,90,85}    % muted dark text
\definecolor{morandiorange}{RGB}{242,222,196} % soft sand/orange for advanced baselines
\usepackage{pifont} % 用于 \ding 打勾打叉
\newcommand{\cmark}{\textcolor{green!60!black}{\ding{51}}} % 绿色的勾
\newcommand{\xmark}{\textcolor{red}{\ding{55}}} % 红色的叉
\newcommand{\samethanks}[1][\value{footnote}]{\footnotemark[#1]}

\definecolor{boxgrey}{RGB}{245,245,245}
\definecolor{linegrey}{RGB}{200,200,200}
\definecolor{boxred}{RGB}{253,240,240}
\definecolor{linered}{RGB}{230,180,180}
\definecolor{boxpurple}{RGB}{248,242,252}
\definecolor{linepurple}{RGB}{210,190,230}
\definecolor{boxorange}{RGB}{255,246,238}
\definecolor{lineorange}{RGB}{240,200,160}
\definecolor{boxcyan}{RGB}{238,248,250}
\definecolor{linecyan}{RGB}{180,220,230}
\definecolor{boxgreen}{RGB}{240,250,242}
\definecolor{linegreen}{RGB}{180,220,190}
\begin{document}

\title{CM2: Multimodal Cultural Reasoning\\ via an Integrated Multi-Agent Framework}
\titlerunning{CM2: Multimodal Cultural Reasoning}
% === Double-blind: author and institution info suppressed for review ===
% \author{Anonymous Author(s)}
% \authorrunning{Anonymous Author(s)}
% \institute{}
\author{%
Qi Li\inst{1,2}\fnmsep\thanks{These authors contributed equally to this work.} \and
Zhaojie Kang\inst{2,3}\fnmsep\samethanks \and
Yingjie He\inst{2} \and
Zheng Lin\inst{2} \and
Hao Zhang\inst{2} \and
\\Guangxin Wu\inst{2} \and
Yan Gong\inst{2} \and
Rong Fu\inst{2} \and
Jianyuan Ni\inst{4}
% \,\Envelope
}
\authorrunning{Q. Li, Z. Kang et al.}
\institute{Lanzhou University, Lanzhou, China \and
Peking University, Beijing, China \and
Taiyuan University of Technology, Taiyuan, China \and
Independent Researcher\\
\email{liq23@lzu.edu.cn, kangzj980@outlook.com}}
\maketitle              % typeset the header of the contribution

% =========================
% Abstract
% =========================

\begin{abstract}
Multimodal Large Language Models (MLLMs) have shown remarkable success in STEM domains, where progress is often driven by vertical, step-by-step deduction under relatively stable symbol systems. Their horizontal, interdisciplinary cultural reasoning, however, remains underexplored.
We propose \textbf{CM2}, a multi-agent framework grounded in the cognitive pathway of human cultural interpretation. CM2 integrates multimodal perception, retrieval-augmented generation, networked reasoning, gated fusion, and reward-driven feedback.
Experiments on CM2D across multiple MLLM backbones show consistent gains over CoT and typical reasoning paradigms; ablations validate each module's contribution, and conflict analyses confirm genuine cross-modal arbitration.

\end{abstract}

% =========================
% 1. Introduction
% =========================
\section{Introduction}
Multimodal Large Language Models (MLLMs)~\cite{Qwen2.5-VL} have rapidly evolved into general-purpose assistants capable of reasoning over mixed visual--textual inputs. Their strongest gains are observed in structured STEM benchmarks~\cite{kang2026quanteval}, which typically emphasize \emph{vertical reasoning}: domain-internal, step-by-step deduction under relatively stable symbolic rules and well-defined objectives.

In contrast, cultural understanding tasks---such as interpreting artworks, artifacts, or historically situated imagery---require a fundamentally different regime. Building on the humanities and social sciences (HSS) perspective articulated in HSSBench~\cite{kang2026hssbench}, we characterize this regime as \emph{horizontal reasoning}: integrating heterogeneous evidence across modalities and disciplines while resolving ambiguity among multiple plausible interpretations~\cite{wei2023chainofthoughtpromptingelicitsreasoning}. Cultural reasoning is inherently context-sensitive: the same motif may convey distinct meanings across periods, styles, or traditions, making evidence attribution and conflict resolution central rather than optional.

Despite its importance, current cultural evaluations remain limited. Many tasks reduce interpretation to simplified QA or recognition-style decisions~\cite{gao2026laobench}, under-testing interdisciplinary integration and ambiguity handling. As a result, MLLMs often produce fluent yet weakly grounded explanations, over-trust superficial visual cues, or generate culturally plausible but unsupported narratives~\cite{yu2026seeing}.

\begin{figure}[!t]
    \centering
    \includegraphics[width=\linewidth]{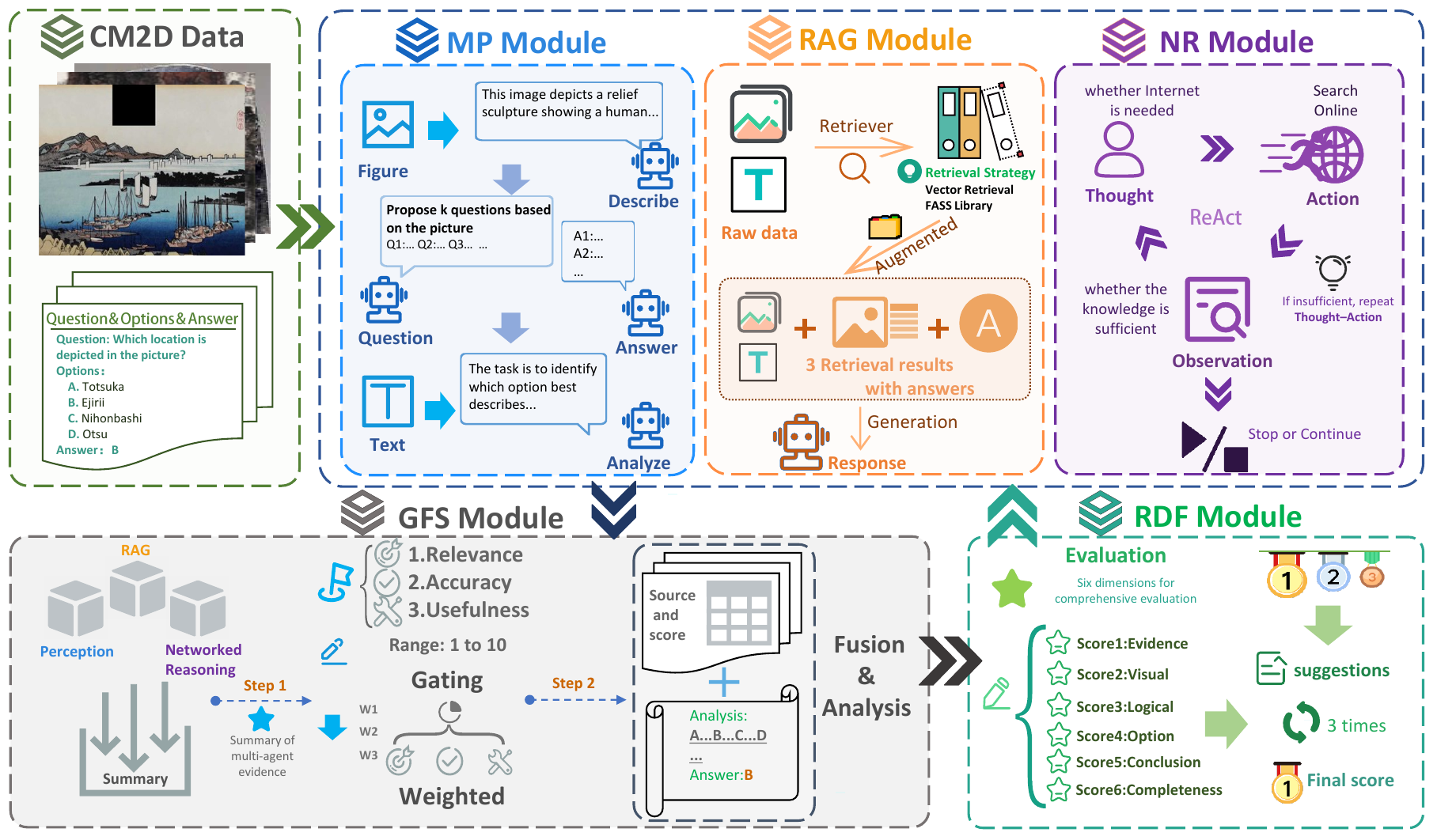}
    \caption{Overview of CM2. The framework orchestrates multimodal perception (MP), retrieval-augmented generation (RAG), networked reasoning (NR), gated fusion and synthesis (GFS), and reward-driven feedback (RDF) for iterative cultural reasoning.}
    \label{fig:placeholder}
\end{figure}

To address these challenges, we propose \textbf{CM2}, an integrated multi-agent framework for structured cultural reasoning, grounded in the cognitive pathway of human aesthetic appreciation~\cite{leder2014ten}. CM2 produces complementary evidence from perception, retrieval, and external verification, arbitrates cross-modal conflicts through gated fusion, and iteratively refines outputs under cultural reasoning criteria.

Complementing the framework, we construct \textbf{CM2D}, a curated multimodal cultural dataset probing horizontal reasoning rather than surface recognition, built via a hybrid expert--agent pipeline with 442 expert-verified test samples and 3,715 retrieval entries.

Our contributions are:
\begin{itemize}[leftmargin=1.2em,itemsep=0.2em,topsep=0.2em]
\item \textbf{Framework.} We introduce CM2, a five-module framework that explicitly targets horizontal, cross-modal cultural reasoning through parallel evidence generation, conflict-aware fusion, and feedback-driven refinement.
\item \textbf{Dataset.} We develop CM2D, a curated multimodal cultural dataset constructed via an expert--agent pipeline, emphasizing context-sensitive interpretation and multimodal dependency.
\item \textbf{Evaluation.} Experiments show consistent gains over CoT and typical reasoning baselines across multiple backbone MLLMs; ablations and conflict analyses confirm the gains stem from genuine cross-modal arbitration.

\end{itemize}

% =========================
% 2. Related Work
% =========================
\section{Related Work}\label{sec:related}
\subsection{MLLMs and Multimodal Reasoning}
Recent MLLMs have achieved strong performance in multimodal instruction following and reasoning~\cite{Qwen2.5-VL}. Benchmarks such as multimodal math and STEM reasoning~\cite{kang2026quanteval} have driven progress in visual encoders, scaling, and prompting. However, these settings predominantly test \emph{vertical} deduction under fixed symbol systems. Cultural interpretation instead requires integrating heterogeneous cues, handling ambiguity, and reasoning under context shifts. This gap motivates specialized frameworks that go beyond generic prompting or retrieval~\cite{lewis2021retrievalaugmentedgenerationknowledgeintensivenlp,zheng2026should,meng2026group,jiang2026agent,kang2026multimodal,qian2026penny}.

\subsection{Cultural and HSS-Oriented Evaluation}
Cultural evaluation remains comparatively underdeveloped. Existing efforts often simplify cultural tasks into direct QA or recognition-like decisions~\cite{gao2026laobench}, which may miss interpretive nuance. HSSBench~\cite{kang2026hssbench} provides a complementary diagnostic perspective by systematically characterizing how large models struggle with humanities and social sciences tasks: shallow symbolic associations, ungrounded narratives, and weak evidence reconciliation. While HSSBench is primarily language-centric, its failure modes become more severe in multimodal cultural settings, where evidence can be distributed across style, iconography, and historical context. CM2 is designed as a direct response: producing structured evidence, resolving conflicts, and refining outputs under explicit cultural criteria.

% =========================
% 3. Method
% =========================
\section{Method}\label{sec:methods}

\subsection{Cognitive Foundations of Horizontal Reasoning}

We characterize cultural interpretation as \emph{horizontal} reasoning: unlike the \emph{vertical}, monotonic deduction of STEM, it demands actively integrating and arbitrating heterogeneous evidence that often conflicts across modalities and contexts. This regime mirrors human aesthetic cognition. Leder and Nadal's~\cite{leder2014ten} five-stage model describes interpretation not as linear deduction but as a dynamic cycle of perceptual analysis, schema activation, classification, cognitive mastering (i.e., conflict arbitration), and evaluation. Drawing on this architecture, CM2 maps each cognitive stage to a dedicated module (Figure~\ref{fig:placeholder}): multimodal perception (MP) extracts stylistic cues; retrieval-augmented generation (RAG) activates cultural schemas; networked reasoning (NR) verifies uncertain claims externally; gated fusion (GFS) arbitrates conflicting evidence via source-specific weighting; and reward-driven feedback (RDF) evaluates and refines the output under culturally grounded criteria.

% ------------------------------------------------
\subsection{Task Formulation}

We formalize a task as $\mathcal{T}=(I,Q,O,A_{\mathrm{gt}})$, where $I$ is an image, $Q$ a query, $O=\{o_1,\ldots,o_K\}$ a candidate option set, and $A_{\mathrm{gt}}$ is the ground-truth answer. CM2 generates three evidence streams, namely perceptual ($E_{\mathrm{mp}}$), retrieved ($E_{\mathrm{rag}}$), and verified external ($E_{\mathrm{nr}}$), which are fused and refined into a final prediction:
\begin{equation}
(A^*, R^*) = \mathrm{RDF}\big(\mathrm{GFS}(E_{\mathrm{mp}},E_{\mathrm{rag}},E_{\mathrm{nr}}), \mathcal{T}\big).
\end{equation}

% ------------------------------------------------
\subsection{Multimodal Perception (MP)}

MP uses the backbone VLM to extract culturally salient perceptual cues $E_{\mathrm{mp}}=\{D_{\mathrm{mp}},\mathcal{Q}\mathcal{A}_{\mathrm{mp}},S_{\mathrm{mp}}\}$: $D_{\mathrm{mp}}$ describes style, composition, and culturally relevant visual attributes (e.g., brushwork, color palette, spatial organization); $\mathcal{Q}\mathcal{A}_{\mathrm{mp}}$ is a structured QA block probing motif identity, likely period, and region of origin; and $S_{\mathrm{mp}}$ is a one-sentence summary of the task-relevant semantic focus.

% ------------------------------------------------
\subsection{Retrieval-Augmented Generation (RAG)}

RAG anchors interpretation by retrieving culturally relevant precedents from a curated knowledge base ($\mathrm{KB}_{\mathrm{rag}}$, 3,715 image--text entries; Sec.~\ref{sec:dataset}). Visual similarity $\mathrm{sim}_{\mathrm{vis}}$ uses CLIP embeddings and textual similarity $\mathrm{sim}_{\mathrm{text}}$ uses a sentence-transformer encoder, both L2-normalized cosine similarities FAISS-indexed; they are combined with $\alpha{=}0.5$ as:
\begin{equation}
\mathrm{sim}(\mathcal{T}, \mathcal{T}_j) = \alpha\cdot\mathrm{sim}_{\mathrm{vis}} + (1-\alpha)\cdot\mathrm{sim}_{\mathrm{text}},
\end{equation}
and the top-$K{=}5$ entries are concatenated as structured context for downstream fusion.

% ------------------------------------------------
\subsection{Networked Reasoning (NR)}

When symbolic claims from MP and RAG remain uncertain, NR performs iterative external verification via web search~\cite{yu2026seeing}. At each round $t$, the MLLM converts the unresolved claim into a query $q^{(t)}$, issues it to a search API, and accumulates the top-$r{=}5$ snippets into a buffer $h^{(t)}$ (initialized $h^{(0)}=\emptyset$):
\begin{align}
q^{(t)} &= f_{\mathrm{query}}(\mathcal{T}, h^{(t-1)}), \\
h^{(t)} &= \mathrm{concat}\big(h^{(t-1)}, \mathrm{Search}(q^{(t)})\big).
\end{align}
The loop runs for at most $T_{\max}{=}3$ iterations, and all accumulated evidence is summarized into a verification report $E_{\mathrm{nr}}=f_{\mathrm{summ}}(h^{(T)})$.

% ------------------------------------------------
\subsection{Gated Fusion \& Synthesis (GFS)}

Because the three sources may conflict, GFS performs relevance-weighted arbitration rather than naive aggregation~\cite{zheng2025medcoact}. For each source $x\in\{\mathrm{mp},\mathrm{rag},\mathrm{nr}\}$, the backbone VLM scores the evidence on direct relevance, factual accuracy, and usefulness for discriminating options (each 1--10), averaged into a raw score $g_x$. Scores are normalized by softmax and used to weight the source embeddings $\mathbf{e}_x$:
\begin{align}
\tilde{g}_x &= \frac{\exp(g_x / \tau_g)}{\sum_{x'}\exp(g_{x'} / \tau_g)}, \\
\mathbf{e}_{\mathrm{fused}} &= \sum_x \tilde{g}_x\, \mathbf{e}_x,
\end{align}
with $\tau_g{=}0.7$ controlling the sharpness of weighting. A synthesis prompt $f_{\mathrm{synth}}$ then generates the fused answer $A_{\mathrm{gfs}}$ and rationale $R_{\mathrm{gfs}}$ from the three evidence texts (with their gate weights) and the task, so that culturally grounded evidence can override superficial cues when conflicts arise.

% ------------------------------------------------
\subsection{Reward-Driven Feedback (RDF)}

RDF evaluates the fused output against six criteria---evidence integration (multiple sources cited and reconciled), visual grounding (anchored in specific visual details), logical flow, option differentiation (each option distinctly assessed), conclusion strength, and completeness (required output structure)---each scored 1--10 by the backbone VLM and averaged into $\bar{s}$~\cite{wang2026beyond,shi2026spader}. If $\bar{s}<\tau{=}0.6$, the VLM summarizes the weakest dimensions into targeted feedback, injected into the next NR and GFS iteration; after at most two rounds, the answer with the highest $\bar{s}$ is selected ($A^*=\arg\max_t\bar{s}^{(t)}$). Unlike generic self-refinement, RDF pinpoints which cultural dimension is weak.

\section{Dataset}\label{sec:dataset}

We introduce \textbf{CM2D}, a multimodal cultural dataset designed to evaluate \emph{horizontal} and context-sensitive reasoning beyond surface recognition. CM2D contains \textbf{4,157} image--question pairs: \textbf{442} expert-verified test samples and \textbf{3,715} retrieval entries ($\mathrm{KB}_{\mathrm{rag}}$). Each instance is a 4-way multiple-choice question with one correct answer and three culturally plausible distractors, covering fine-grained reasoning skills such as symbolism, style, period/region attribution, medium conventions, and contextual interpretation.

\begin{wrapfigure}{r}{0.45\linewidth}
    \centering
    \includegraphics[width=\linewidth]{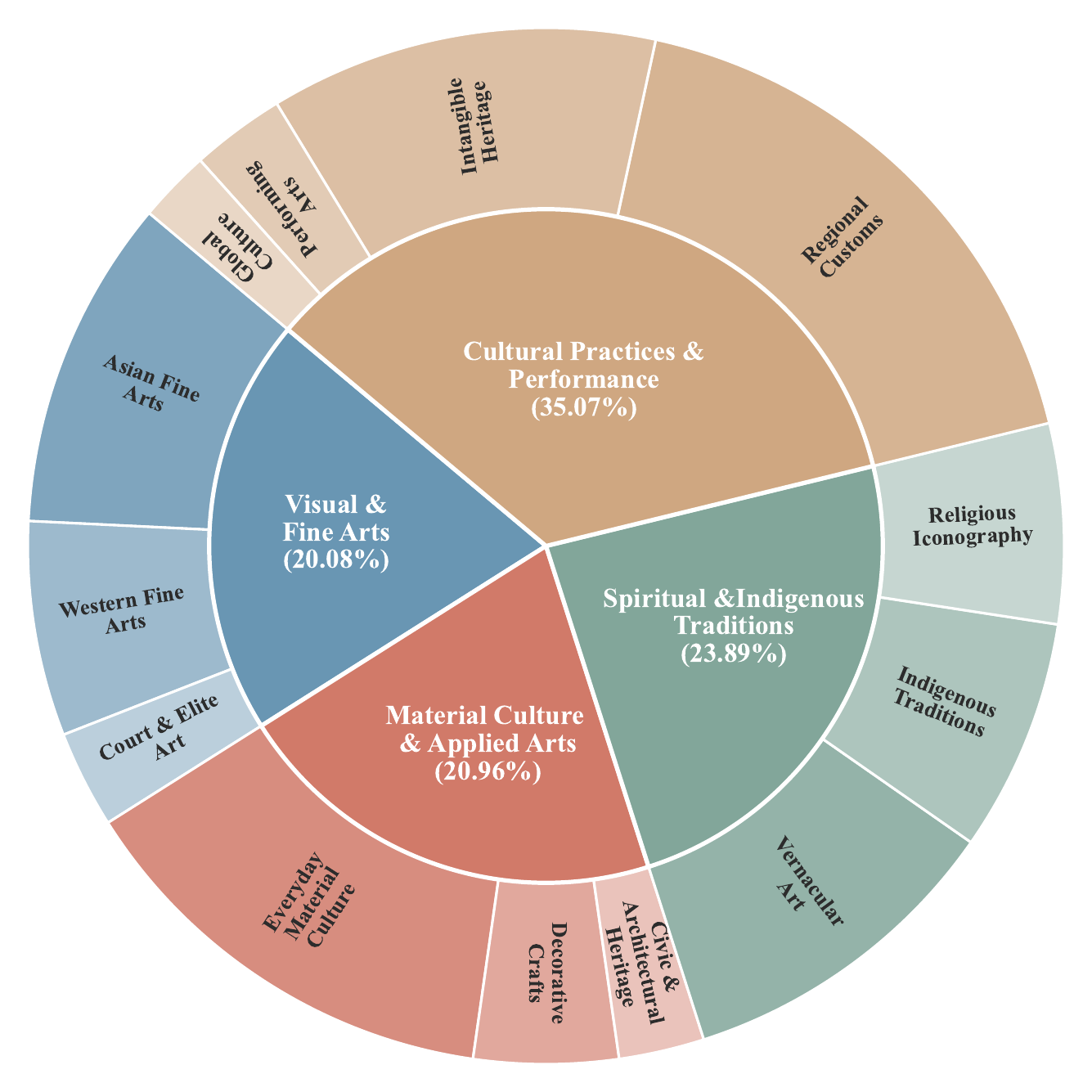}
    \caption{Taxonomy of CM2D: four coarse categories (inner ring) and thirteen fine-grained sub-categories (outer ring).}
    \label{fig:taxonomy}
\end{wrapfigure}

\paragraph{Taxonomy and Visual Diversity.}
To counter Eurocentric bias and ensure diverse visual representation, CM2D is structured into four coarse categories and thirteen fine-grained taxonomies (Figure~\ref{fig:taxonomy}). This categorization is principally based on the primary cultural function and social medium of each artifact. Acknowledging the inherent intersectionality of the humanities (e.g., a religious fresco serving as both fine art and spiritual iconography), we assign the most salient label to each instance to facilitate structured evaluation. The four core categories are:

\setlength{\intextsep}{0pt}

(i) Visual \& Fine Arts: High-aesthetic value creations including Asian Fine Arts, Western Fine Arts, and Court \& Elite Art.

(ii) Material Culture \& Applied Arts: Objects with physical spatial and utilitarian presence, covering Everyday Material Culture, Decorative Crafts, and Civic \& Architectural Heritage.

(iii) Spiritual \& Indigenous Traditions: Items reflecting religious or grassroots beliefs, including Religious Iconography, Vernacular Art, and Indigenous Traditions.

(iv) Cultural Practices \& Performance: Dynamic human-centric traditions, spanning Regional Customs, Intangible Heritage, Performing Arts, and Global Culture.

\paragraph{Hybrid Construction Pipeline.}
Data are sourced via a rigorous hybrid expert--agent pipeline to ensure both diversity and depth. This pipeline operates in three phases. \emph{Phase I: Multimodal Aggregation.} Based on expert-provided seeds, the agent retrieves authoritative materials (e.g., verified museum repositories, open-source textbooks) to gather high-information-density text and images. \emph{Phase II: Semantic Drafting.} Specialized LLM agents parse the textual background to extract key symbolic attributes and dynamically generate culturally plausible distractors. \emph{Phase III: Dual-Tier Validation.} All drafted items undergo automated redundancy checks followed by meticulous human expert review, where experts refine skill tags and approve the final answer keys.

\paragraph{Multimodal dependency check.}
To discourage unimodal shortcuts, we apply screening under two counterfactual settings:
(i) \textbf{text-only}, removing the image while keeping $(Q,O)$; and
(ii) \textbf{image-only}, keeping $I$ while replacing $Q$ with a generic prompt and shuffling $O$~\cite{li2025consistent}.
Items that remain trivially solvable under either setting are revised (e.g., tightening distractors or removing leaked cues) or discarded. A final expert pass confirms that decisive evidence requires cross-modal grounding rather than pure memorization or object spotting.

\paragraph{Dataset Composition and Quality Assurance.}
We intentionally prioritize data quality and expert-verified density over sheer scale. Within the 3,715 $\mathrm{KB}_{\mathrm{rag}}$ entries, approximately 40\% are directly expert-authored, while 60\% are agent-drafted and subsequently expert-verified. Conversely, to ensure the highest level of academic rigor and eliminate noisy evaluations, the 442-sample test set is 100\% human-curated and cross-validated by independent experts. During the validation phase, any candidate question lacking genuine cultural ambiguity was discarded.

% =========================
% 5. Experiments
% =========================
\section{Experiments}\label{sec:exp}
We evaluate CM2 on CM2D and report 4-way multiple-choice accuracy on the \textbf{442}-sample expert-verified test set, using Qwen2.5-VL-3B-Instruct, Qwen2.5-VL-7B-Instruct~\cite{Qwen2.5-VL}, llava-onevision-7b~\cite{li2024llavaonevisioneasyvisualtask}, and GPT4.1-mini~\cite{openai2025gpt41} as backbones. As a reference, Baseline performs single-pass Chain-of-Thought (CoT) prompting~\cite{wei2023chainofthoughtpromptingelicitsreasoning} with greedy decoding and no retrieval, external search, or iterative refinement. Beyond CoT, we further compare three typical reasoning paradigms under the same backbone and identical test set: Self-Refine~\cite{madaan2023selfrefineiterativerefinementselffeedback} (two critique-and-revise rounds), CoT-SC~\cite{wang2023selfconsistency} ($N{=}5$ chains with temperature $0.7$ and majority voting), and Agent Debate~\cite{du2023improving} (two proposer--critique rounds). \textbf{CM2} uses the same backbone across all modules with deterministic decoding (temperature $0$) and the hyperparameters in Sec.~\ref{sec:methods}; prompts and budgets are held fixed across backbones for comparability.

\subsection{Main Results}
Table~\ref{tab:main_results} reports the main accuracy results on CM2D. CM2 yields consistent improvements across all backbones, suggesting that horizontal cultural reasoning benefits from explicit evidence integration and conflict arbitration rather than scaling alone. Notably, llava-onevision-7b more than doubles accuracy (23.53\% $\rightarrow$ 50.68\%), while the stronger GPT4.1-mini still improves (41.63\% $\rightarrow$ 55.20\%), indicating that gains are systematic rather than limited to weaker models.

Crucially, CM2 also maintains a significant edge over typical reasoning paradigms. Taking llava-onevision-7b as an example, CoT-SC and Agent Debate achieve 40.50\% and 43.21\% respectively, notably lagging behind CM2's 50.68\%. This validates our core hypothesis: generic multi-hop reasoning or unguided debate often struggles with horizontal cultural tasks, as they lack the specific mechanisms to resolve explicit cross-modal conflicts via external grounding. Across backbones, CM2 maintains a clear margin over the best baseline (4.75--10.86pp), narrowing to 0.90pp on the strongest backbone GPT4.1-mini. 

\setlength{\belowcaptionskip}{0pt}  % 图注贴近表格线
\setlength{\intextsep}{4pt}         % 表格与上下正文的外间距
\begin{table}[!t]
\centering
\footnotesize
\setlength{\tabcolsep}{2pt}
\renewcommand{\arraystretch}{0.88}
\caption{Accuracy (\%) on the CM2D test set. The best result in each row is \textbf{bold}; the best baseline (excluding CM2) is \underline{underlined}.}
\label{tab:main_results}
\begin{tabular}{lccccc}
\toprule
Model & Baseline (CoT) & Self-Refine & CoT-SC & Agent Debate & CM2 \\
\midrule
Qwen2.5-VL-3B-Instruct & 23.98 & 37.56 & 38.91 & \underline{39.59} & \textbf{44.34} \\
Qwen2.5-VL-7B-Instruct & 29.41 & 35.52 & \underline{35.97} & 33.26 & \textbf{46.83} \\
llava-onevision-7b       & 23.53 & 34.40 & 40.50 & \underline{43.21} & \textbf{50.68} \\
Qwen2.5-VL-7B (SFT)     & 38.46 & --      & --      & --      & \textbf{47.06} \\
GPT4.1-mini              & 41.63 & 49.10 & \underline{54.30} & 54.07 & \textbf{55.20} \\
\bottomrule
\end{tabular}
\end{table}

\subsection{Ablation Studies}
To understand which components drive the improvements, we ablate each module while keeping all other components unchanged. Table~\ref{tab:ablation} shows that every module provides complementary value. MP contributes fine-grained stylistic and compositional cues; RAG provides precedent grounding for iconography and culturally conventional symbols; NR reduces historically plausible but unsupported narratives; GFS is crucial when evidence sources disagree; and RDF improves targeted correction when the initial synthesis is weak. 

\setlength{\belowcaptionskip}{0pt}  % 图注贴近表格线
\setlength{\intextsep}{4pt}         % 表格与上下正文的外间距
\begin{table}[!t]
\centering
\setlength{\tabcolsep}{3.5pt}
\renewcommand{\arraystretch}{0.88}
\caption{Ablation results (\%) on the CM2D test set. The drop from removing each module quantifies its contribution; the largest drop per column is \underline{underlined}.}
\label{tab:ablation}
\begin{tabular}{lccc}
\toprule
Ablation & Qwen2.5-VL-7B & llava-onevision-7b & Qwen2.5-VL-3B \\
\midrule
CM2 (full) & 46.83 & 50.68 & 44.34 \\
\midrule
w/o MP      & \underline{37.78} & 44.12 & 42.99 \\
w/o RAG     & 40.50 & \underline{39.59} & \underline{39.37} \\
w/o NR      & 42.99 & 42.53 & 40.27 \\
w/o GFS     & 40.27 & 39.82 & 43.21 \\
w/o RDF     & 45.02 & 44.57 & 43.21 \\
\bottomrule
\end{tabular}
\end{table}

\subsection{Fine-Grained Evaluation under Evidence Conflict}\label{sec:mechanism}

While the ablation study (Sec.~\ref{sec:exp}) confirms the necessity of each module, this section investigates the behavioral patterns of CM2 when faced with cross-modal evidence conflict. By comparing the independent predictions of MP, RAG, and NR before fusion, we dynamically partition the test set into two mutually exclusive subsets: the \emph{Consensus Subset} (where all three sources agree) and the \emph{Conflict Subset} (where at least one source disagrees). We analyze these subsets on two representative backbones (Qwen2.5-VL-7B and llava-onevision-7b).

\begin{figure}[!t]
    \centering
    \includegraphics[width=0.8\linewidth]{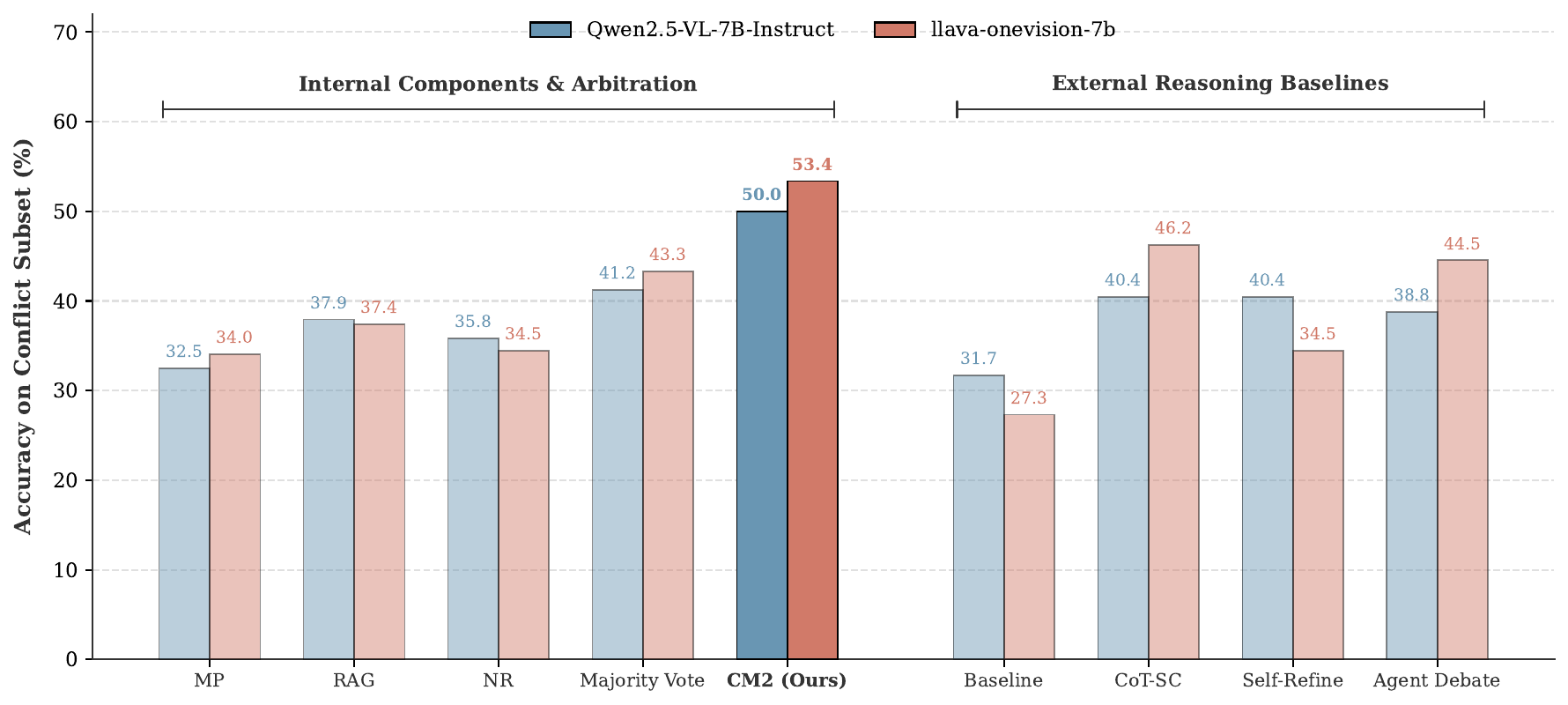}
    \caption{Accuracy on the Conflict Subset:CM2's gated arbitration substantially outperforms Majority Voting, CoT-SC, and Agent Debate.}
    \label{fig:mechanism}
\end{figure}

\paragraph{Consensus as a strong baseline.} 
Before addressing conflicts, we note that on the complementary Consensus Subset (45.7\%--46.2\% of samples), CM2 already leads across both backbones (+4.9--11.4pp over the best baseline). This confirms that the enrichment from MP and RAG provides consistent, foundational benefits even when evidence is perfectly aligned.

\paragraph{Conflict is pervasive and challenging.} 
The core arbitration challenge resides in the Conflict Subset, which constitutes the majority of test samples (54.3\% for Qwen2.5-VL-7B-Instruct and 53.8\% for llava-onevision-7b), confirming that evidence-source disagreement is pervasive in cultural tasks. On this challenging subset, the single-pass CoT baseline scores only 31.7\% (Qwen) and 27.3\% (llava). A simple Majority Vote among the pre-fusion agents improves this to 41.2\% and 43.3\%, while external baselines (CoT-SC: 40.4\%, 46.2\%; Agent Debate: 38.8\%, 44.5\%) offer only incremental gains. In contrast, CM2’s relevance-weighted gated fusion reaches \textbf{50.0\%} and \textbf{53.4\%}---a substantial $+18.3$--$26.1$pp over the CoT baseline and $+7.2$--$9.9$pp over the best competing method (Figure~\ref{fig:mechanism}). This demonstrates that passive aggregation, sampling, and debate each leave a substantial gap: only explicit retrieval grounding with conflict-aware gating effectively resolves cross-modal evidence conflict.

\setlength{\intextsep}{8pt}         % 表格与上下正文的外间距
% === NEWLY MODIFIED START ===

\begin{figure}[!t]
\centering
\setlength{\fboxsep}{3pt}
\footnotesize

% ---------- 1. Task Info (Grey) ----------
\fcolorbox{linegrey}{boxgrey}{%
\begin{minipage}[t]{\linewidth}
\vspace{0pt}

\textbf{\textcolor{morandidark}{Case: Lotus Form (CM2D)}}\hfill
\textcolor{morandidark}{\textbf{Type:} Symbolic \& stylistic}
\vspace{2pt}\hrule\vspace{2pt}

\begin{minipage}[t]{0.28\linewidth}
\vspace{0pt}\centering
\fcolorbox{linegrey}{white}{\includegraphics[width=0.70\linewidth]{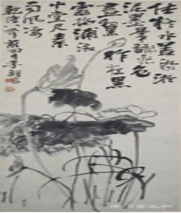}}
\end{minipage}
\hfill
\begin{minipage}[t]{0.70\linewidth}
\vspace{0pt}
\textbf{\textcolor{morandidark}{Question:}}\\
\textcolor{morandidark}{What characteristics are shown in the form of the lotus flowers in the artwork?}\\[0.4em]
\textbf{\textcolor{morandidark}{Options:}}\\
\textcolor{morandidark}{A. Vibrant vitality and growth} \hfill \cmark \\
\textcolor{morandidark}{B. The withered state of autumn} \hfill \xmark \\
\textcolor{morandidark}{C. A sense of etherealness filled with Eastern philosophy} \hfill \xmark \\
\textcolor{morandidark}{D. The desolation of winter} \hfill \xmark
\end{minipage}

\vspace{0pt}
\end{minipage}
}

% ---------- 2. Baselines (consolidated into one box) ----------
\fcolorbox{linered}{boxred}{%
\begin{minipage}[t]{\linewidth}
\vspace{0pt}
\textbf{\textcolor{morandidark}{$\blacklozenge$ Baseline (CoT): B}} \xmark \quad
\textit{Reasoning:} The lotus flowers and leaves point downwards, painted with heavy, dark ink strokes. \underline{This drooping shape indicates decay} and autumn's withered state. Thus, it depicts the end of a life cycle rather than growth.

\vspace{2pt}
\textbf{\textcolor{morandidark}{$\blacklozenge$ Self-Refine: C}} \xmark \quad
\textit{Reasoning:} \textit{[Critique]} The initial thought is too literal. \textit{[Refinement]} Eastern ink paintings prioritize \underline{philosophical emptiness} over physical seasons; drooping lines express etherealness.

\vspace{2pt}
\textbf{\textcolor{morandidark}{$\blacklozenge$ CoT-SC (N=5): B}} \xmark \quad
\textit{Reasoning:} B(3), C(1), D(1). Despite temperature sampling, all paths are \underline{misled by the superficial cue} of bent stems, yielding a confident but incorrect consensus on autumnal decay.

\vspace{2pt}
\textbf{\textcolor{morandidark}{$\blacklozenge$ Agent Debate: B}} \xmark \quad
\textit{Reasoning:} \textit{Agent A} argues dark ink shows \underline{autumn decay (B)}; \textit{Agent B} argues barren background suggests \underline{winter desolation (D)}. Without external cultural anchors, they default to superficial decay cues and choose B.
\end{minipage}
}

% ---------- 6. CM2 (Green) ----------
\fcolorbox{linegreen}{boxgreen}{%
\begin{minipage}[t]{\linewidth}
\textbf{\textcolor{morandidark}{$\blacklozenge$ CM2 (Ours): A}} \cmark \quad
\textbf{\textit{Arbitrated Reasoning Process:}}
\begin{itemize}[leftmargin=1.0em,itemsep=0.1em,topsep=0.1em,parsep=0pt,partopsep=0pt]
\item \textbf{MP:} Stems show upward tension despite downward-facing leaves.
\item \textbf{RAG:} Retrieved critiques define this ``drooping but unbroken'' style as inner vigor.
\item \textbf{NR:} External sources confirm the lotus symbolizes resilience in this artist's context.
\item \textbf{GFS:} \underline{Drooping is an expressive technique, not literal decay.} Grounded evidence overrides visual bias, confirming vitality.
\end{itemize}
\end{minipage}
}

\caption{\textbf{Qualitative case:} Baselines overfit visual cues or hallucinate stereotypes, while CM2 arbitrates evidence to recover vitality (A). Reasoning texts are human-edited summaries, not verbatim model outputs.}
\label{fig:lotus_case}

\end{figure}

\subsection{Qualitative Case Study}

Figure~\ref{fig:lotus_case} presents a qualitative case study that instantiates common HSS-style failure modes. While typical reasoning paradigms (CoT, Self-Refine, CoT-SC, Agent Debate) either overfit to superficial visual cues (drooping implies decay) or hallucinate generic stereotypes (Eastern emptiness), CM2 effectively integrates perceptual style with retrieved cultural knowledge to accurately derive the underlying symbolic meaning (resilience and vitality).

% =========================
% 6. Conclusion
% =========================
\section{Conclusion}
We presented CM2, a multi-agent framework that targets horizontal cultural reasoning through parallel evidence generation, retrieval grounding, external verification, gated conflict arbitration, and reward-driven refinement. We introduced CM2D, a curated multimodal dataset emphasizing context-sensitive interpretation with multimodal dependency. Across multiple backbones, CM2 improves substantially over CoT and typical reasoning paradigms (Self-Refine, CoT-SC, Agent Debate), with ablations confirming each module's contribution and conflict analyses validating genuine cross-modal arbitration.

\bibliographystyle{splncs04}
\bibliography{references}

\end{document}